\documentclass[
final
]{ceurart}

\usepackage{listings}
\usepackage{overpic}
\usepackage{wrapfig}
\usepackage{subcaption}
\usepackage{graphicx}
\usepackage{amsmath}
\usepackage{amsfonts}
\usepackage{amsthm}
\usepackage{xspace}

\newcommand{\ALT}{{\textbf{ALTERNATE}}\xspace}
\newcommand{\SWI}{{\textbf{SWITCH}}\xspace}
\newcommand{\PRO}{{\textbf{TEXTUAL}}\xspace}
\newcommand{\UNE}{{\textbf{UNET}}\xspace}

\definecolor{ourred}{RGB}{252,0,0}
\definecolor{ourblue}{RGB}{0,0,252}
\newcommand{\ourred}{\color{ourred}}
\newcommand{\ourblue}{\color{ourblue}}

\begin{document}

\copyrightyear{2024}
\copyrightclause{Copyright for this paper by its authors.
	Use permitted under Creative Commons License Attribution 4.0
	International (CC BY 4.0).}

\conference{}

\title{How to Blend Concepts in Diffusion Models}

\author{Lorenzo Olearo}[%
	orcid=0009-0000-7290-3549,
	email=lorenzo.olearo@unimib.it,
	url=https://lorenzo.olearo.com/,
]
\fnmark[1]

\author{Giorgio Longari}[%
	orcid=0000-0002-2086-9091,
	email=giorgio.longari@unimib.it,
]
\fnmark[1]

\author{Simone Melzi}[%
	orcid=0000-0003-2790-9591,
	email=simone.melzi@unimib.it,
	url=https://sites.google.com/site/melzismn/,
]
\author{Alessandro Raganato}[%
	orcid=0000-0002-7018-7515,
	email=alessandro.raganato@unimib.it,
	url=https://raganato.github.io/,
]
\author{Rafael Pe\~naloza}[%
	orcid=0000-0002-2693-5790,
	email=rafael.penaloza@unimib.it,
	url=https://rpenalozan.github.io/,
]
\address{University of Milano-Bicocca, Milan, Italy}

\fntext[1]{These authors contributed equally.}

\begin{abstract}
	For the last decade, there has been a push to use multi-dimensional (latent)
	spaces to represent concepts; and yet how to manipulate these concepts or reason
	with them remains largely unclear. Some recent methods exploit multiple latent
	representations and their connection, making this research question even more
	entangled. Our goal is to understand how operations in the latent space affect
	the underlying concepts. To that end, we explore the task of concept blending
	through diffusion models. Diffusion models are based on a connection between a
	latent representation of textual prompts and a latent space that enables image
	reconstruction and generation. This task allows us to try different text-based
	combination strategies, and evaluate easily through a visual analysis. Our
	conclusion is that concept blending through space manipulation is possible,
	although the best strategy depends on the context of the blend.
\end{abstract}


\maketitle

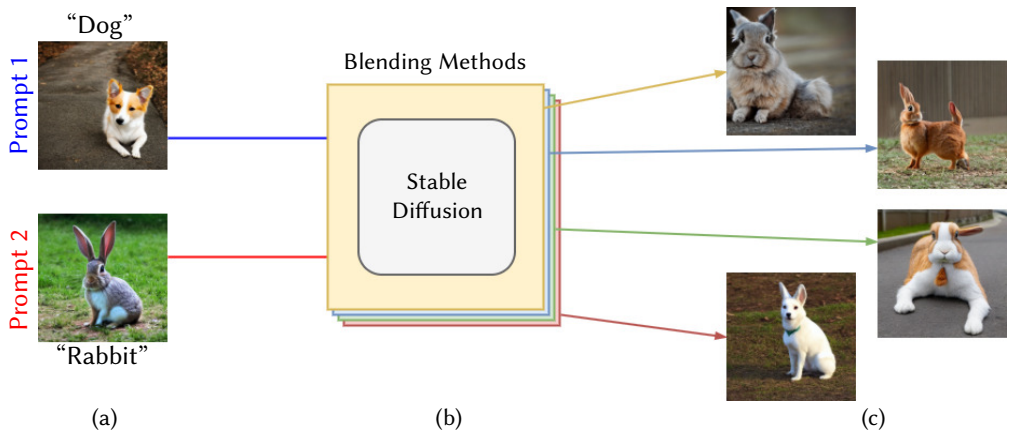
\begin{figure}[h!]
\centering
        \begin{overpic}[width=0.8\textwidth]{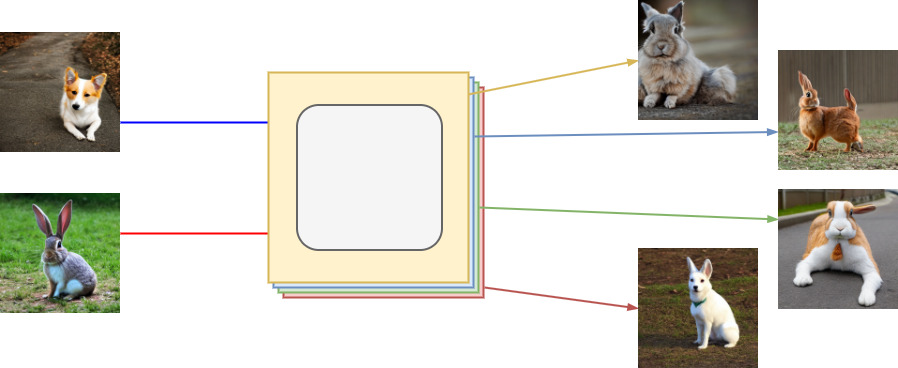}
         \put(-3, 25.6){\ourblue \rotatebox{90}{Prompt 1}}
         \put( 2.9, 38.3){``Dog''}
         \put( 1.8, 3.8){``Rabbit''}
         \put(-3,7.3){\ourred \rotatebox{90}{Prompt 2}}
          \put(31.5,34.5){\footnotesize Blending Methods}
          \put(38,22){\footnotesize Stable}
          \put(36.5,19){\footnotesize Diffusion}
        \put(5.5,-2.5){\footnotesize (a)}
        \put(41.0,-2.5){\footnotesize (b)}
        \put(85.0,-2.5){\footnotesize (c)}
    
    \end{overpic}
    \vspace{0.36cm}
    
    \captionof{figure}{A visualization of the proposed analysis. From left to right: (a) given two input textual concepts (``dog'', ``rabbit''), (b) four different techniques are applied to explore multiple ways to blend them together through stable diffusion and (c) the obtained outputs are compared with qualitative analysis and a user study. }
\end{figure}

\section{Introduction}
\label{sec:introduction}

The field of knowledge representation deals with the task of, indeed, 
\emph{representing} the knowledge of a domain in a manner that can be used 
for intelligent applications~\cite{BrLe-04KR}. Over the decades, most of the
progress in the field has focused on logic-based knowledge representation 
languages, and their reasoning capabilities. In this setting, concepts---the
first-class citizens of any domain representation---are formalised by limiting
the interpretations that they can be assigned to, and their connections with 
other concepts.

A different, more implicit approach has been attempted for the last decade. 
This approach represents concepts as points (or sometimes volumes) in
a multidimensional so-called \emph{latent} space. This representation (also known as the
\emph{embedding}) is built considering the semantic similarities and
differences between concepts. Although at an abstract level this representation
is similar to \citeauthor{Gard04}'s \emph{conceptual spaces}~\cite{Gard04}, 
there are essential differences between the two approaches; most notably, concept
composition cannot be achieved through simple set operations. More generally, 
although the use of the latent space is becoming more common, it remains unclear how
to navigate it and, hence, how to reason within this representation.

Our overarching goal is to understand the properties of the latent space and,
in particular, how different operations within it affect the underlying concepts.
It is usually understood that every point in the latent space represents a 
concept, and thus, navigating it has the potential of creating new concepts.
In this paper, we focus on the question of \emph{concept blending}; briefly, the task of 
creating a new concept which combines (``blends'') the properties of two or more
concepts~\cite{FaTu-08} (see Section~\ref{sec:preliminaries} for more details). 
We explore the possibility of constructing such blends
through (text-to-image) diffusion models starting from textual prompts descriving the concepts. This choice is motivated by, first,
the easy access to the latent space through the textual prompts and, second, the
ability to evaluate the quality of the results visually.

We study different strategies for concept blending which exploit the overall
architecture of Stable Diffusion~\cite{rombach2022high}. 
None of these methods relies on further training or fine-tuning,
but they all focus on the topology of the latent space and the architecture of Stable Diffusion.
We notice that,
in general, visual depictions of concept blends can be automatically generated
through these techniques, although their quality may vary. An empirical study
was used to evaluate the relative performance of each method. The results 
suggest that there is no absolute best method, but the choice of blending
approach depends on the combined concepts.

Notably, the task of concept blending considered here is just a milestone
towards our general goal of understanding the properties of the latent space 
and how navigating it affects the underlying concepts. This understanding will
be useful towards an explainable use of latent spaces and embeddings in general.

\section{Preliminaries}
\label{sec:preliminaries}

Human creativity has always been the key to the innovating process, giving us the possibility to 
imagine things which are yet to be discovered, and diverging scenarios to explore. In recent years, the field of artificial 
intelligence~(AI) has been revolutionized by generative models, which 
are capable of creating new and original contents by exploiting the countless examples these models 
have been trained on.
Among the multiple variants and possibilities to exploit generative AI, diffusion models 
like Stable Diffusion~\cite{podell2023sdxl}, Dall-E, or Midjourney produce as output original images 
based on textual prompts or images given as input for the model.
To provide clarity for this work, we introduce the fundamental concepts and 
components of Stable Diffusion and the notions of concept blending.

\subsection{Stable Diffusion}  

\setlength{\columnsep}{5pt}
\begin{wrapfigure}[8]{r}{0.5\linewidth}
    \centering
    \vspace{-0.6cm}

    \begin{overpic}[width=\linewidth] 
    {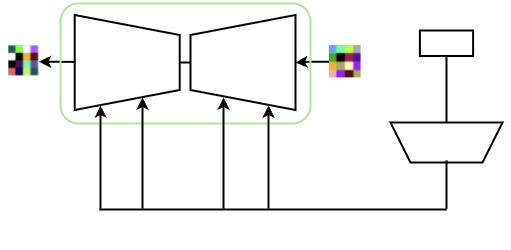}
    
    \put(19.6,45.4){\textbf{\footnotesize Denoising U-Net}
     }
    \put(46.0,35.5){\footnotesize $E$}
    \put(22.6,35.5){\footnotesize $D$}  
    \put(74.6,44.9){\footnotesize \textbf{input prompt}}
    \put(86.5,39.5){\footnotesize $p$}
    \put(81.9,22.2){\footnotesize prompt}
    \put(81.0,18){\footnotesize encoder}     

    \end{overpic}
\end{wrapfigure}

Stable Diffusion (SD) is a text-to-image generative model developed by
\citeauthor{rombach2022high} in 2022~\cite{rombach2022high}, which follows the
typical architecture of diffusion models \cite{pmlr-v37-sohl-dickstein15}
comprising a forward and a backward process. In the forward process, a clean
sample from the data (in this case, an image) is sequentially corrupted by
random noise reaching, after a defined number of steps, pure random noise. In
the backward process, a neural network is trained to sequentially remove the
noise, thereby restoring the clean data distribution; this is the main phase
intervening during image generation. The Stable Diffusion network architecture
utilized during the backward phase is principally made up of (i) a Variational
Autoencoder (VAE) \cite{kingma2013auto}, (ii) a U-Net \cite{ronneberger2015u},
and (iii) an optional text encoder. The VAE characterizes SD as a Latent
Diffusion Model, mapping images into a lower-dimensional space through an
\emph{encoder}, followed by a diffusion model to craft the distribution over
encoded images. The images are then represented as points in the latent space
($\mathbb{R}^n$). Afterwards, a \emph{decoder} is needed to convert a point back
into an image.

The U-Net is composed of an encoder-decoder pair, where the bottleneck contains
the compact embedding representation of the images. The encoder $E$ maps the
input samples according to the given prompt embedding into this latent
embedding, then the decoder $D$ processes this latent embedding together with its prompt
embedding to reconstruct a sample that is as close as possible to the original
one. The U-Net and text embedding are crucial in conditioning the output
generated by the model. At each step of the denoising process, the prompt
embedding is injected into the three blocks of the U-Net via cross-attention
mechanism. In this way, the textual prompt conditions the denoising process and
in turn the generation of an image. The prompt embedding is generated by the
text encoder, following the pipeline of SD 1.4. In our experiments, as text
encoder, we adopt a pre-trained CLIP Vit-L/14.


With these details we can now establish the focus of this study, summarizing it 
in the research question: \textit{can diffusion models produce visual blends
of two concepts?} Identifying each concept through a word, we want to create a 
new image that simultaneously represents a combination of both, simulating the 
human capacity for associative thinking. 
To address this problem, we present various methodologies leveraging SD as the 
backbone of our experiments. But before, we explain the notion of concept blending.

\subsection{Concept Blending}

\emph{Blending} represents a cognitive mechanism that has been innately 
exploited to create new abstractions from familiar concepts~\cite{confalonieri2018concept}. This 
process is often experienced in our daily interactions, even during a 
casual conversation. This conceptual framework has been studied over 
the past three decades~\cite{costello2000efficient}, 
offering a model that incorporates mental spaces and conceptual 
projection. It examines the dynamic formation of interconnected 
domains as discourse unfolds, aiming to discover fundamental 
principles that underlie the intricate logic of everyday 
interactions. In this context, a mental space is a temporary 
knowledge structure, which is dynamically created, for the purpose of 
local understanding, during a social interaction. It is composed of 
elements (concepts) and their interconnections. It is context-
dependent and not necessarily a description of 
reality~\cite{fauconnier1994mental}. 
This general notion can be specified in different notions. For 
our purpose, we are interested in \emph{visual conceptual blending}, 
which combines aspects of conceptual blending and visual blending.

\emph{Conceptual Blending} is the operation that constructs a partial 
match between two or more input mental spaces, and project them into 
a new ``blended'' one \cite{fauconnier1994mental}. This 
blended space has common characteristics of the input spaces, 
allowing a mapping between its elements and their counterparts in 
each input space. However, it also generates a new emergent 
conceptual structure, which is unpredictable from the input spaces 
and not originally present in them. Therefore, blending occurs at the 
conceptual level, yet representations of these blended concepts are 
valuable and frequently employed in 
advertising~\cite{joy2009conceptual} and other 
domains~\cite{kutz2014pluribus}.

The \emph{Visual Blending} process, instead, is essential to generate 
new visual representations, such as images, through the fusion of at 
least two existing ones. There are two primary options for visual 
blending, according to the style of rendering employed: 
photo-realistic rendering and non-photo-realistic techniques, such as 
drawings. Approaches that focus on text-to-image generation have as 
main goal the visual representation of concepts, and, in the case of 
blending, the topology could to be summarized as a bunch of 
visual operations, as analyzed by 
\citeauthor{phillips2004beyond}~\cite{phillips2004beyond}. One of 
these operators, called \emph{fusion}, partially depicts and merges 
the different inputs to create a hybrid image, allowing for a higher 
coherence between the object parts of the object(s), and helping the 
viewers in perceiving the hybrid object as a unified whole. 
In \emph{replacement}, one input concept is present and its sole 
function is to occupy the usual environment of the other concept, or 
have its shape adapted to resemble the other input. 
\emph{Juxtaposition} is a technique that involves placing two 
different elements side by side, to create a harmonious or provoking 
whole. 
A good example of Visual Blending along with different approaches to the
operations descrived above (and others) can be found
in~\cite{xiao2015vismantic}. It is important to note that high-quality blends
between concepts require that only \emph{some} of the main characteristics of
the input concepts are taken into account~\cite{chilton2021visifit}. Exploiting
the three main visual properties of color, silhouette, and internal details,
helps the creator to obtain a great resulting blend. An image result from
blending can be evaluated by taking in account the number of dimensions (or
visual properties) over which the blend has been applied. 

\medskip


\emph{Visual Conceptual Blending} introduces a model for creating 
visually blended images rounded in strong conceptual reasoning. The 
work by \citeauthor{inproceedings}~\cite{inproceedings} argues that 
visual conceptual blending goes beyond simply merging two images; 
instead, it emphasizes the importance of conceptual reasoning as the 
foundation of the blending process, resulting in both images and 
accompanying conceptual elaborations. These blends have context, are 
grounded in justifications, and can be named independently of the 
original concepts. This contrasts with standard Visual Blending, 
which focuses solely on image merging, and typically involves mapping 
concepts to objects and integrating them while maintaining 
recognisability and inference association.

\medskip
After all these preliminaries, we can rephrase our research question 
in a more specific form as: \emph{can Stable Diffusion models merge two 
semantically distant concepts into a new image, practically 
performing a Visual (Conceptual) Blending operation?} In this study, we investigate the efficacy of diffusion models, which are supposed to recreate each image that should be imagined, in generating high-quality blended images. We assess existing approaches to perform blending with stable diffusion, and propose novel methods. 
To the best of our knowledge, this is the first investigation that evaluates the performance of different blending techniques with diffusion models using only textual prompts. We initially operate on the latent space where the textual prompts are embedded, and then explore alternative methods by directly manipulating the specific
architecture of the diffusion model; more precisely, the U-Net conditioning phase is manipulated to edit the textual prompt that is injected (see Section~\ref{sec:backgorund} for details). To evaluate the results of the methods, we conducted a user survey where the
subjects were asked to rank the outcomes of different blending tasks,
divided in multiple categories.

\section{Blending Methods with Stable Diffusion}
\label{sec:backgorund}

In this section, we briefly review some of the existing approaches for blending
concepts with diffusion models. Some of these methods were already published in
previous work~\cite{melzi2023does}, while others are available on public
implementations, but without a full description of their details.  
We mention explicitly whenever we are unsure if our implementation matches exactly the one proposed in the reference.

\paragraph{Experimental setup}
We fix the generative network $\mathcal{G}$ as Stable Diffusion v1.4
\cite{rombach2022high} with the UniPCMultistepScheduler \cite{zhao2023unipc} set
at 25 steps. This version uses a fixed pretrained text encoder (CLIP
Vit-L/14~\cite{radford2021learning}).  All the images are generated as 512x512
pixels with the diffusion process carried in FP16 precision in a space
downscaled by a factor of 4. The conditioning signal is provided only in the
form of textual prompts, and the guidance scale is set to 7.5.

In our study we focused on Stable Diffusion as a good trade-off between quality
and computational cost; however, the blending methods that we analyzed can be
implemented in other diffusion models with no latent downscale.
Our entire implementation of the blending methods in their respective pipelines
together with some of the generated samples is open source and available on our
GitHub repository.%
\footnote{Project repository: \url{https://github.com/LorenzoOlearo/blending-diffusion-models}}

An important feature of many generative methods, which allows
them to produce varying outputs on the same prompt, is the use
of a pseudo-random number generator (and pseudo-random noise)
which can be established through a seed.
Given an input textual prompt $p$, and a seed $s$, we denote as 
$I_{s,p} =\mathcal{G}(s,p)$ the image generated by the model 
$\mathcal{G}$ given the input prompt $p$ and the seed $s$. 
Prompts will be usually denoted with the letter $p$, sometimes
with additional indices to distinguish between them; e.g., 
$p_1$ and $p_2$ when two different prompts are used simultaneously.
Given a prompt $p$, $p^*$ denotes its \emph{latent representation};
that is, the multi-dimensional vector obtained from the
encoding operation. Similarly, $p^*_1$ and $p^*_2$ denote
the latent representations of $p_1$ and $p_2$, respectively.

\subsection{Blending in the Prompt Latent Space (\PRO)} 
\setlength{\columnsep}{5pt}
\begin{wrapfigure}[8]{r}{0.5\linewidth}
    \centering
    \vspace{-0.6cm}

    \begin{overpic}[width=\linewidth] 
    {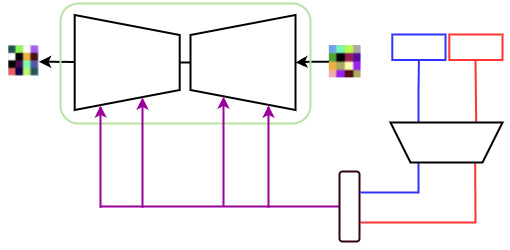}
    \put(19.6,45.4){      \textbf{\footnotesize Denoising U-Net}
     }
    \put(46.0,35.5){\footnotesize $E$}
    \put(22.6,35.5){\footnotesize $D$}  
    \put(74.6,44.9){\footnotesize \textbf{input prompts}}
    \put(80.5,39.1){\footnotesize $p_1$}
    \put(91.8,39.1){\footnotesize $p_2$}
    \put(81.9,22.2){\footnotesize prompt}
    \put(81.0,18){\footnotesize encoder}     

     \put(67.65,4.2){\rotatebox{90}{\footnotesize mean}}

    \end{overpic}
    \label{fig:schemaPROMPT}
\end{wrapfigure}
The first method examined in this study is the one
recently proposed by \citeauthor{melzi2023does}~\cite{melzi2023does}. This approach exploits the relationship between conceptual blending and vector operations within the prompt latent space.

As depicted in the inset figure, given the two input prompts $p_1$ and $p_2$, we first compute their latent representations $p^{*}_1$ and $p^{*}_2$ through the prompt encoder. Then, we define the blended latent vector as the Euclidean mean between $p^{*}_1$ and $p^{*}_2$.
Finally, we generate the blended image by conditioning the Stable diffusion model with the estimated blended latent vector.

It is essential to underline that blending in the latent space representing the prompts does not correspond to
blending directly the images, as in a visual blending process. Instead, it means generating an image representing a specific fusion of the concepts provided
as the input textual prompts.
Indeed, the Euclidean mean between the two representations
is a (potentially unexplored) point of the latent space which
intuitively represents the concept that is closest to both
input concepts, thus defining an ``in-between'' characterisation.

Although in this paper we only consider the mean of the two latent representations of the input prompts, we highlight that \citeauthor{melzi2023does} consider also other linear combinations of $p^{*}_1$ and $p^{*}_2$ to avoid fully symmetric constructions. 
A similar technique was implemented in the \emph{Compel open source library},%
\footnote{Compel: \url{https://github.com/damian0815/compel}}
which performs the weighted blend of two textual prompts.

\setlength{\columnsep}{5pt}
\begin{wrapfigure}[8]{r}{0.5\linewidth}
    \centering
    \vspace{-0.6cm}

    \begin{overpic}[width=\linewidth] 
        {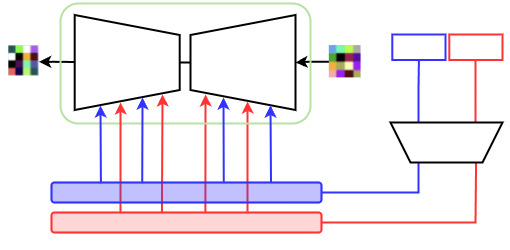}

        \put(19.6,45.4){\textbf{\footnotesize Denoising U-Net}
        }
        \put(46.0,35.5){\footnotesize $E$}
        \put(22.6,35.5){\footnotesize $D$}
        \put(74.6,44.9){\footnotesize \textbf{input prompts}}
        \put(80.5,39.1){\footnotesize $p_1$}
        \put(91.8,39.1){\footnotesize $p_2$}
        \put(81.9,22.2){\footnotesize prompt}
        \put(81.0,18){\footnotesize encoder}

        \put(9.8,10.2){
        \footnotesize \ourblue{The first N denoising iterations}}
        \put(11.6,4.4){\footnotesize \ourred{The last M denoising iterations}}
        \end{overpic}
    \end{wrapfigure}

    \subsection{Prompt Switching in the Iterative Diffusion Process (\SWI)}
    \label{sec:switch}

    This blending technique involves switching the textual prompt during the
    iterative process of the diffusion model. The inference process first starts
    with a single prompt $p_1$ and then, at a certain iteration, the prompt is
    switched to $p_2$ until the end of the diffusion process. The generation is thus
    conditioned on both prompts leading to an image that, when the switch is
    executed at the right timestep, blends the two concepts. Intuitively, \SWI{}
    starts by generating the general shape of $p_1$, but then fills out the details
    based on $p_2$ thus producing a visual blend of the two concepts.

    It is crucial to choose the right iteration to switch the prompt. Unfortunately,
    this is an intrinsic challenge for each new image and does not depend only on
    the geometric distance between the $p_1^*$ and $p_2^*$ embeddings. From our
    experiments, it was observed that the optimal iteration for this switch is
    directly related to the spatial similarity between the image generated by the
    model conditioned only on $p_1$ and the one generated by $p_2$.

    This technique has also been implemented into the \emph{Stable Diffusion web UI}
    developed by \emph{AUTOMATIC1111}.%
    \footnote{Stable Diffusion web UI: \url{https://github.com/AUTOMATIC1111/stable-diffusion-webui}}
    Among its numerous functionalities, this implementation allows prompt editing
    during the mid-generation of an image.



    \setlength{\columnsep}{5pt}
    \begin{wrapfigure}[8]{r}{0.5\linewidth}
        \centering
        \vspace{-0.6cm}

        \begin{overpic}[width=\linewidth]
        {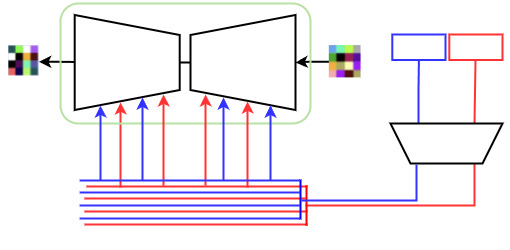}
    \put(19.6,45.4){\textbf{\footnotesize Denoising U-Net}
     }
    \put(46.0,35.5){\footnotesize $E$}
    \put(22.6,35.5){\footnotesize $D$}  
    \put(74.6,44.9){\footnotesize \textbf{input prompts}}
    \put(80.5,39.1){\footnotesize $p_1$}
    \put(91.8,39.1){\footnotesize $p_2$}
    \put(81.9,22.2){\footnotesize prompt}
    \put(81.0,18){\footnotesize encoder}  
    
    \put(63,13.7){
    \footnotesize \ourblue{Even}}
    \put(60.6,10.7){
    \footnotesize \ourblue{iteration}}
      \put(63.5,5.4){
    \footnotesize \ourred{Odd}}
    \put(60.6,2.4){
    \footnotesize \ourred{iteration}}
    \end{overpic}
\end{wrapfigure}

\subsection{Alternating Prompts in the Iterative Diffusion Process (\ALT)}
Recall that in general diffusion models, at each timestep defined by the
scheduler of the diffusion process, the noise in the sample is estimated by the
U-Net model. This estimation is performed by the model with knowledge both of
the timestep and the conditioning signal (i.e., the prompt). 

The Alternating Prompt technique consists of conditioning the U-Net with a
different prompt at each timestep. The first prompt $p_1$ is shown to the U-Net
at even timesteps, while the second prompt $p_2$ is shown at odd timesteps. By
performing this alternating prompt technique, the diffusion pipeline can
successfully generate an image that blends the two given prompts. Even though at
different timesteps, the U-Net is conditioned by both prompts during the
diffusion process. Moreover, the blending ratio can be controlled by adjusting
the number of iterations in which each prompt is shown to the U-Net.

One can intuitively think of this approach as an alternating superposition of
the generation process between $p_1$ and $p_2$. This method is also implemented
in the \emph{Stable Diffusion web UI} developed by \emph{AUTOMATIC1111}.

\section{Method}
\label{sec:method}
In this section, we propose a different blending paradigm to visually combine
two textual prompts in the diffusion pipeline. In a standard diffusion
architecture, given a single input prompt $p$, its corresponding embedding $p^*$
is injected with a cross-attention mechanism in the three main blocks of the
U-Net: the encoder, the bottleneck, and the decoder. During the encoding and
bottleneck steps, the $p^*$ embedding is used to guide the compression of the
input sample into a latent representation that accurately maps the concept $p$
that is being generated. Then, during the decoding phase, the $p^*$ embedding is
used to guide the reconstruction of the sample towards the distribution of the
concept $p$ that is being generated. Our idea arises from this compression and
reconstruction operation and it is described in the following subsection. To the
best of our knowledge, this method has not been proposed before.

\subsection*{Different Prompts in Encoder and Decoder Components of the U-Net (\UNE)}
\setlength{\columnsep}{5pt}
\begin{wrapfigure}[8]{r}{0.5\linewidth}
    \centering
    \vspace{-0.5cm}

    \begin{overpic}[width=\linewidth] 
            {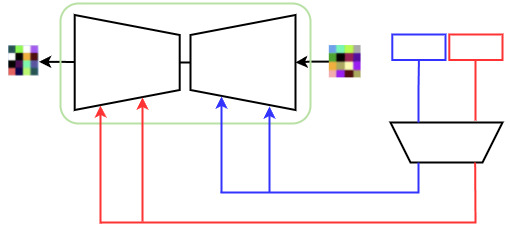}
    \put(19.6,45.4){\textbf{\footnotesize Denoising U-Net}
     }
    \put(46.0,35.5){\footnotesize $E$}
    \put(22.6,35.5){\footnotesize $D$}  
    \put(74.6,44.9){\footnotesize \textbf{input prompts}}
    \put(80.5,39.1){\footnotesize $p_1$}
    \put(91.8,39.1){\footnotesize $p_2$}
    \put(81.9,22.2){\footnotesize prompt}
    \put(81.0,18){\footnotesize encoder}  
    \end{overpic}
\end{wrapfigure}
We implement our new method only using text-based conditioning, but the method
can theoretically be extended to other conditioning domains.

As we briefly describe above, the U-Net architecture is composed of three main
blocks: the encoder, the bottleneck and the decoder. Each of these block
receives the prompt embedding $p^*$ as input together with the sample from which
the noise has to be estimated. 

The key idea our method involves guiding the compression of the sample into the
bottleneck block with a first prompt embedding $p_1^*$. Then, guide its
reconstruction towards the distribution of the second prompt $p_2$ by injecting
into the decoder block the embedding $p_2^*$ as visualized in the inset Figure.
This allows the U-Net to construct a latent representation for the sample
matching the concept described by $p_1$ and then reconstruct the sample with the
features of the second prompt $p_2$.

The expected result from this technique is to obtain an image that globally
represents or recalls the concept described by $p_1$ and simultaneously shows
some of the features that typically describe the concept of the second prompt
$p_2$.
From our findings, changing the prompt embedding in the bottleneck block does
not significantly affect the final result. Consequently, we keep the prompt
$p_1$ in the decoder and bottleneck blocks while we change the prompt $p_2$ in
the encoder block.

\section{Validation and Results}
\label{sec:results}

\begin{figure}[th!]
    \centering
    \begin{overpic}
        [width=0.7\linewidth]{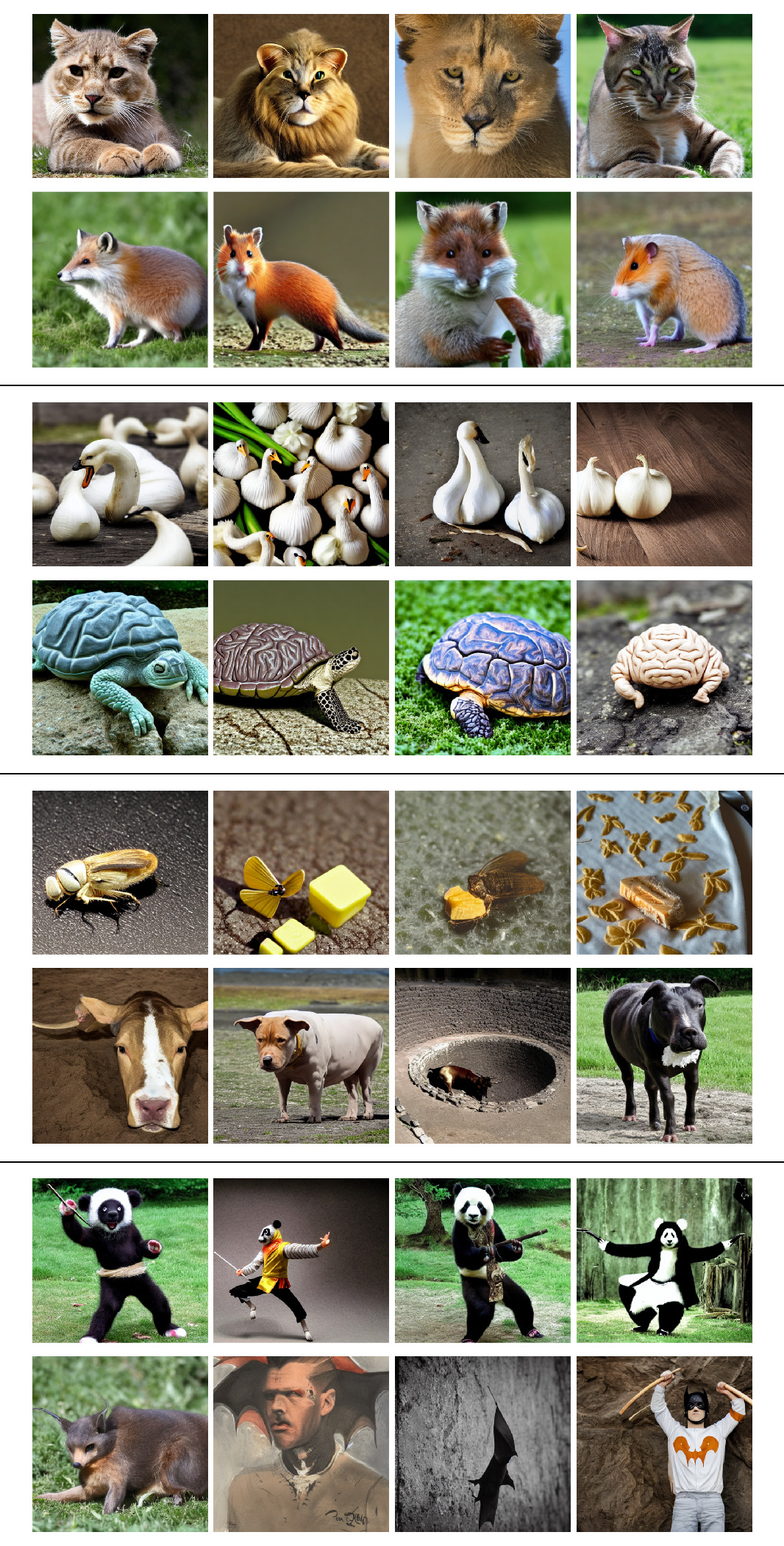}
        \put(4.2, 99.5){{\PRO}}
        \put(16.5, 99.5){{\SWI}}
        \put(26.8, 99.5){{\ALT}}
        \put(41, 99.5){{\UNE}}

        \put(-1.3, 11.4){\rotatebox{90}{Real}}
        \put(-1.3, 32){\rotatebox{90}{Compound Words}}
        \put(-1.3, 58){\rotatebox{90}{Object + Animal}}
        \put(-1.3, 83){\rotatebox{90}{Pair of Animals}}

        \put(0.26, 4){\rotatebox{90}{Man-Bat}}
        \put(0.26, 13.4){\rotatebox{90}{Kung fu-Panda}}
        \put(0.26, 29){\rotatebox{90}{Bull-Pit}}
        \put(0.26, 40){\rotatebox{90}{Butter-Fly}}
        \put(0.26, 53){\rotatebox{90}{Turtle-Brain}}
        \put(0.26, 64.8){\rotatebox{90}{Garlic-Swan}}
        \put(0.26, 77.6){\rotatebox{90}{Fox-Hamster}}
        \put(0.26, 90.8){\rotatebox{90}{Lion-Cat}}
    \end{overpic}
    \caption{Samples of two blend per category.}
    \label{fig:survey-sample-images}
\end{figure}
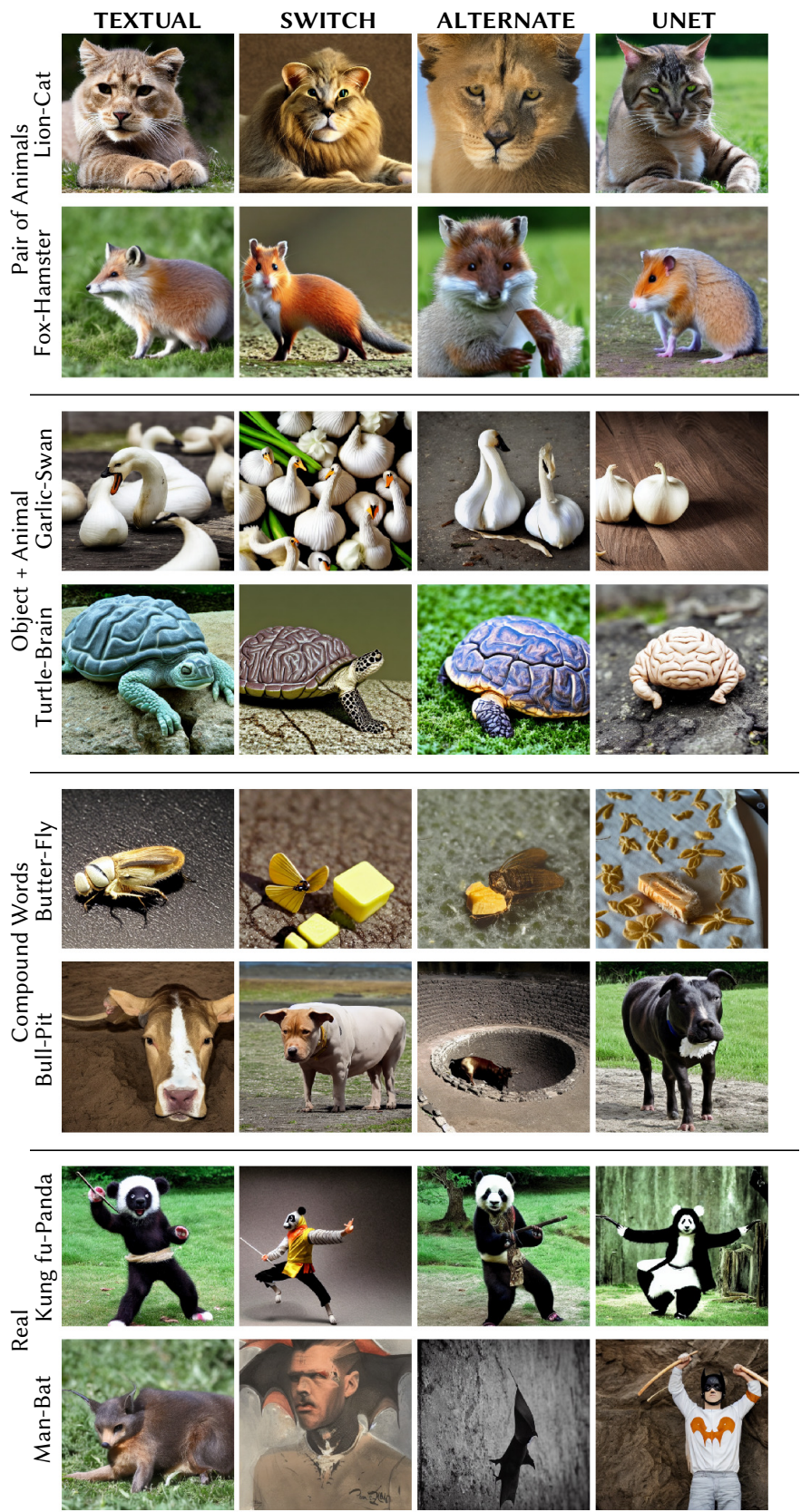

We now describe our experimental setting and analysis made to
evaluate the four blending approaches presented in the previous
sections, applied over two simple conceptual prompts. The outputs
of these models can be visualized in 
Figures~\ref{fig:survey-sample-images} and~\ref{fig:comparison}.

The experiments aimed to assess previously proposed blending methods across four distinct macro-categories, which are visually explained in Figure \ref{fig:survey-sample-images}. 
The four categories are \emph{pair of animals}, \emph{object and animal},
\emph{compound words}, and \emph{real-life scenarios}. These were selected to
showcase different kinds of blending of concepts, which are expected to showcase
diverging properties. For pairs of animals, we expect that the shared
characteristics between the concepts will aid the blending process; the use of
object and animal concepts in the second category is expected to widen the
semantic gap between the input prompts, but produce more ``creative'' artifacts.
The third category delved into objects representing compound words, offering a
more conceptual blending challenge. Here, we observed how the methods responded
to prompts comprising the compound's constituent parts, which are not literal
descriptions of the target object but, rather are interpretable as a figure of
speech or metaphor. We aimed to investigate whether the models should learn the
necessary abstractness to perform a blending similar to the concept associated
with the compound word, or reach a new visual blending that merges the
characteristics of the two prompts. For the last category, drawing inspiration
from real-world visual blend examples, regardless of their underlying concepts,
we derived prompts to condition the models, allowing us to investigate their
adaptability and ability to reconstruct well-known blends.

\subsection*{User Analysis}
    To impartially evaluate the quality of the methods, we conducted a survey involving 23 participants and presented 24 images, arising from the four categories described. The survey was
    constructed as follows. We first selected 24 concept pairs 
    covering examples from the four macro-categories; each concept in the pair was described through a simple one-word prompt (except for ``kung fu''). Then, the four different blending
    methods were used to generate the visual conceptual blend of
    each pair. All images were generated with the same size and
    quality, and presented to the users, with the instruction to
    rank them according to their blending effectiveness (from best
    to worst). 
    Our participant pool was carefully selected to ensure they had no prior experience with blending theory. 

    While the two prompts used to generate the images were provided to the subjects, we deliberately withheld information regarding the model responsible for each image, eliminating potential bias. Additionally, to further mitigate bias, the order of images within each question was randomized for each participant. This approach aimed to discern whether a superior blending method existed among the four proposed and whether certain methods outperformed others within specific categories.
    
    It is essential to note that, for each question, the top four images proposed were selected by the authors from a pool generated using ten different seeds. Given that blending quality across all methods is not entirely independent of seed choice, we aimed to minimize this dependency by carefully selecting the best results. For a better understanding of the evaluation approach, Figure~\ref{fig:survey-sample-images} shows
    some of the images that were presented to the subjects for
    ranking, along with the methods that produced them.
    
\begin{table}[tb]
    \caption{Mean and first-mode rank in the results of the survey over 24 concept pairs (with 23 participants).}
    \label{tab:results}
    \centering
    \begin{tabular}{@{}lc*{4}{>{\columncolor{lightgray!50}}cc}@{}}
        \toprule
         & \multirow{2}{*}{Prompt} & \multicolumn{2}{c}{ALTERNATE} & \multicolumn{2}{c}{SWITCH} & \multicolumn{2}{c}{UNET} & \multicolumn{2}{c}{TEXTUAL} \\
         & & mean & mode & mean & mode & mean & mode & mean & mode \\
         \midrule
         \multirow{6}{*}{\rotatebox{90}{\scalebox{1}{PairOfAnimals}}} 
           & Elephant-Duck & 2.43 & 2 & 3.21 & 4 & 2.65 & 3 & 1.82 & 1 \\
           & Lion-Cat & 2.52 & 2 & 2.30 & 3 & 3.52 & 4 & 1.65 & 1 \\
           & Frog-Lizard & 3.73 & 4 & 1.91 & 2 & 1.41 & 1 & 2.95 & 3 \\
           & Fox-Hamster & 2.68 & 3 & 1.82 & 2 & 3.77 & 4 & 1.82 & 1 \\
           & Rabbit-Dog & 2.95 & 4 & 2.68 & 3 & 2.72 & 2 & 1.68 & 1 \\
           \cmidrule{2-10}
           & CATEGORY TOTAL & 2.86 & 3 & 2.39 & 2 & 2.82 & 4 & 1.98 & 1 \\
        \midrule
         \multirow{8}{*}{\rotatebox{90}{Object+Animal}} 
           & Turtle-Brain & 2.76 & 3 & 1.38 & 1 & 3.10 & 4 & 2.86 & 2 \\
           & Pig-Cactus & 2.38 & 3 & 1.52 & 1 & 3.43 & 4 & 2.76 & 2 \\
           & Garlic-Swan & 1.86 & 1 & 1.81 & 1 & 3.71 & 4 & 2.71 & 2 \\
           & Coconut-Monkey & 1.62 & 2 & 1.67 & 1 & 3.81 & 4 & 2.95 & 3 \\
           & Tortoise-Broccoli &  1.62 & 1 & 3.43 & 4 & 2.48 & 3 & 2.57 & 2 \\
           & Turtle-Wood & 2.19 & 1 & 2.29 & 2 & 2.29 & 3 & 3.33 & 4 \\
           & Turtle-Pizza & 3.52 & 4 & 2.95 & 3 & 1.48 & 1 & 2.10 & 2 \\
           \cmidrule{2-10}
           & CATEGORY TOTAL & 2.28 & 2 & 2.15 & 1 & 2.90 & 4 & 2.76 & 2 \\
        \midrule
         \multirow{10}{*}{\rotatebox{90}{CompoundWords}} 
           & Butter-Fly & 2.57 & 3 & 2.24 & 2 & 2.33 & 1 & 3.00 & 4 \\
           & Dragon-Fly & 2.62 & 3 & 3.43 & 4 & 1.67 & 1 & 2.43 & 2 \\
           & Bull-Pit & 2.81 & 4 & 2.24 & 3 & 2.86 & 3 & 2.23 & 1 \\
           & Blimp-Whale & 2.95 & 3 & 2.05 & 1 & 2.05 & 2 & 3.04 & 4 \\
           & Jelly-Fish & 2.86 & 3 & 1.43 & 1 & 3.14 & 4 & 2.67 & 2 \\
           & Fire-Fighter & 3.00 & 4 & 1.48 & 1 & 2.81 & 2 & 2.90 & 2 \\
           & Tea-Pot & 1.62 & 1 & 2.48 & 2 & 3.81 & 4 & 2.29 & 2 \\
           & Snow-Flake & 3.33 & 4 & 2.62 & 2 & 1.48 & 1 & 2.62 & 3 \\
           & Cup-Cake & 2.71 & 2 & 2.85 & 4 & 2.62 & 4 & 1.90 & 1 \\
           \cmidrule{2-10}
           & CATEGORY TOTAL & 2.72 & 4 & 2.31 & 1 & 2.53 & 4 & 2.57 & 2 \\
        \midrule
         \multirow{4}{*}{\rotatebox{90}{Real}} 
           & Kung fu-Panda & 1.95 & 1 & 2.76 & 3 & 2.05 & 2 & 3.43 & 4 \\
           & Man-Bat & 3.67 & 4 & 1.67 & 1 & 1.67 & 2 & 3.05 & 3 \\
           & Beaver-Duck & 2.24 & 1 & 1.95 & 2 & 3.43 & 4 & 2.48 & 3 \\
           \cmidrule{2-10}
           & CATEGORY TOTAL & 2.62 & 4 & 2.13 & 2 & 2.38 & 2 & 2.98 & 3 \\
        \midrule
        \multicolumn{2}{c}{GLOBAL TOTAL} & 2.61 & 3 & 2.26 & 1 & 2.68 & 4 & 2.54 & 3 \\ 
         \bottomrule
    \end{tabular}
\end{table}

Table~\ref{tab:results} summarises the results of the survey, indicating the
mean and mode (i.e., most frequent) rank given to each method for each prompt
pair and summarizing the results by category and globally. In both cases, a
higher value means a lower quality blend perceived by the subjects of the
survey. The goal of this analysis is to understand which blending method
performs better in general (for the global summary) and in a more fine-grained
manner by category and by prompt pair. We emphasise that the mean value should
be handled with care, as a few low rankings (value 4) can greatly skew to mean
of a ranking that is typically considered of high quality. Indeed, in the last
row of the table we can observe that the average ranking of all methods
throughout the whole experiment is quite similar, even though \SWI is most
frequently selected as the best method, and \UNE as the worst. Worth noticing is
also that the mode does not necessarily provide a full ranking between methods.

In the next section we will discuss the merits of the presented
blending methods and the results of the user survey; yet, for the
moment we can already see that, at least from the perspective of
the ranking given, there is no clear \emph{best} blending approach,
but quality varies between images, and more broadly between 
categories. For instance, \UNE was ranked fourth in three 
categories, but second for the category of real-life scenarios.
Similarly, although \UNE's mode rank in compound words was 4, it was
also the highest ranked in three of the prompt pairs in this 
category.

\section{Discussion}
\label{sec:discussion}

\vspace{12pt}
\begin{figure}[h!]
\centering
    \begin{overpic}[width=0.95\textwidth]{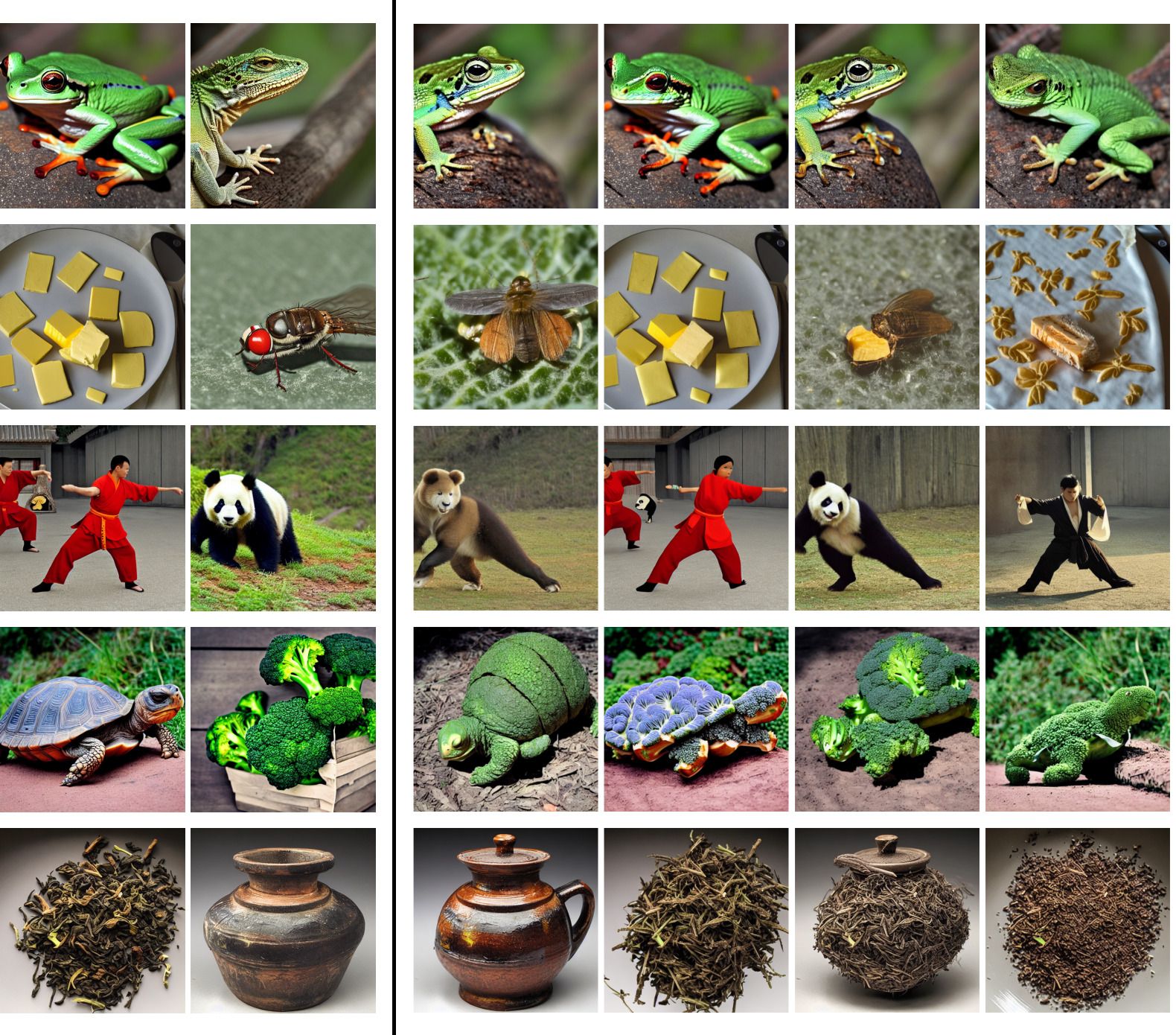}
    \put(-3, 6.3){\rotatebox{90}{\small{Tea-Pot}}}
    \put(-3, 18.5){\rotatebox{90}{\small{Tortoise-Broccoli}}}
    \put(-3, 36.5){\rotatebox{90}{\small{Kung fu-Panda}}}
    \put(-3, 56){\rotatebox{90}{\small{Butter-Fly}}}
    \put(-3, 72.8){\rotatebox{90}{\small{Frog-Lizard}}}

    \put(4, 87.2){\small{prompt 1}}
    \put(19, 87.2){\small{prompt 2}}
    \put(37, 87){\small{\PRO}}
    \put(54, 87){\small{\SWI}}
    \put(68.3, 87){\small{\ALT}}
    \put(88.3, 87){\small{\UNE}}
    \end{overpic}
\caption{
    Comparison of the blending methods. On the left, the individual prompts, and
    on the right, the results of the blending methods. All the images are
    generated starting from the same identical initial noise.
}
\label{fig:comparison}
\end{figure}
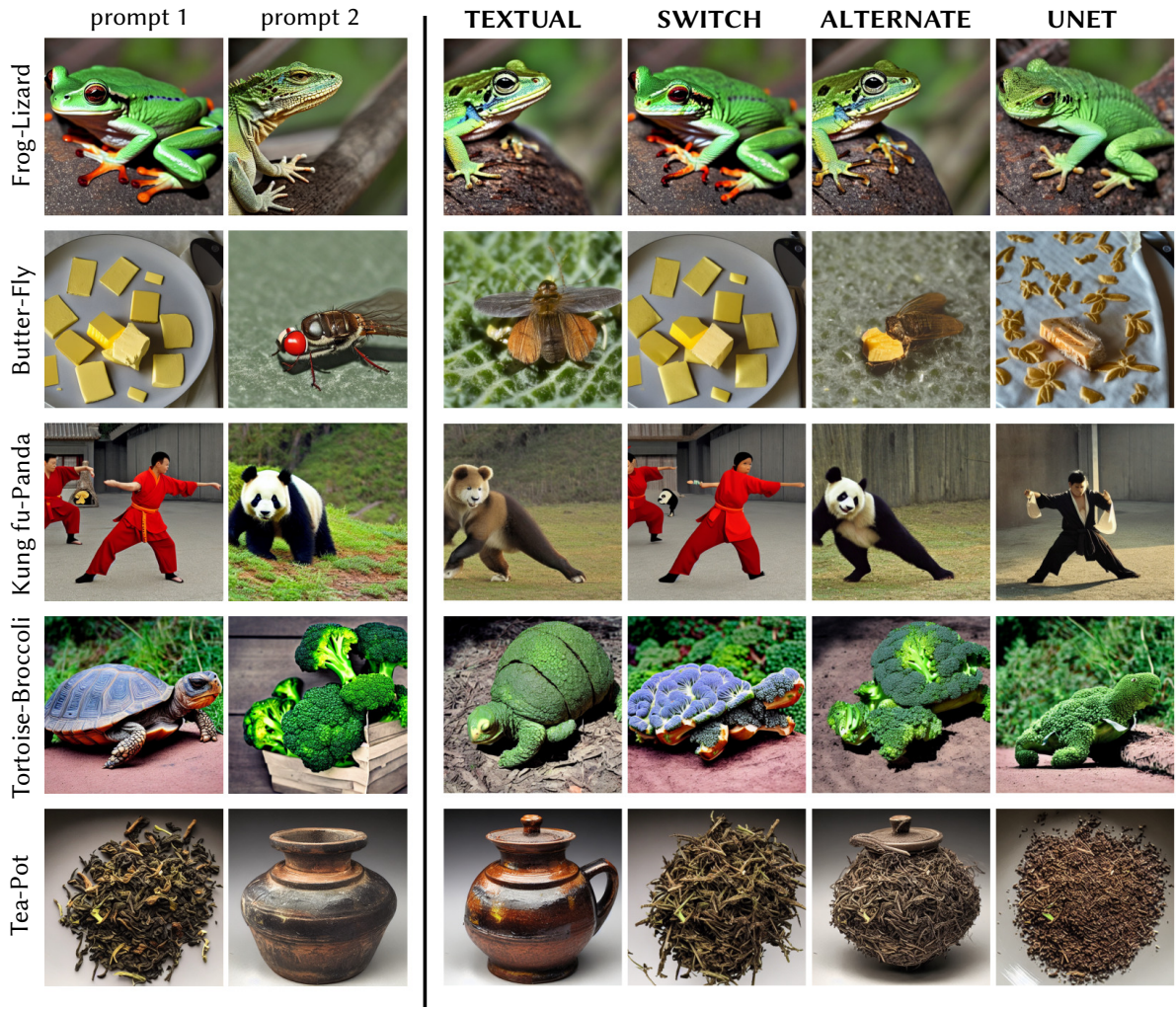

Figure \ref{fig:comparison} shows the results of the four different blending
methods with the prompts \textit{Frog-Lizard}, \textit{Butter-Fly}, \textit{Kung
fu-Panda}, \textit{Tortoise-Broccoli}, and \textit{Tea-Pot}. To better
understand the behavior of each method, all images in each row were generated
using the same seed and thus starting from the same random noise. Moreover, the
blending ratio between the two prompts was kept constant at 0.5 across all
methods.

We measure the \textit{visual distance} of two concepts by visually evaluating
the spatial similarity of the images generated conditioning the pipeline on
them. This is a key aspect to consider when evaluating the quality of the blend
as, with the exception of the \PRO{} which instead focuses on the semantic
blend, it influences the performance of the blending methods.

When it comes to logical blends, one often considers a \textit{main} concept
which is modified by a \textit{secondary} one. That is, the blended concept is
primarily an instance of the main concept, but with some characteristics that
recall the secondary concept. With the exception of \PRO{}, the blending methods
presented in this paper are not symmetric, meaning that the order of the prompts
in the blend affects the final image. This is particularly important when
dealing with compound words like \textit{pitbull}, although this word commonly
refers to a specific breed of dog, its intrinsic semantic and historical meaning
refers to \textit{bull in a pit}. When visually blending the two concepts
\textit{pit} and \textit{bull} with the methods illustrated in this paper, it is
important to take into account which of the two concepts is the \textit{main}
one and which is the \textit{modifier}. 
By analyzing the results in Figure \ref{fig:comparison}, it is evident that this
\textit{primary}-\textit{modifier} relationship is not coherent across all the
analyzed methods. In \PRO{} and \ALT{}, the main concept of the blend appears to
be the second prompt while its modifier is the first one. The opposite behavior
is instead what characterizes \SWI{} and \UNE{} where the main concept of the
blend is the first prompt and the modifier is the second one.

This behavior was not expected and to keep the experiments straightforward all
blends were generated considering the first prompt as the \textit{main}
concept and the second as its \textit{modifier}. This is the reason why when
blending the words that make the compound word \textit{Pitbull}, the blend is
generated as a \textit{Bull}-\textit{Pit} instead of \textit{Pit}-\textit{Bull}.

As expected from the work by \citeauthor{melzi2023does}~\cite{melzi2023does},
performing the blending operation in the latent space of the prompts, as in the
case of \PRO{}, does not always leads to an image that visually blends the two
concepts. This is particularly evident in the case of \textit{Kung fu-Panda},
where the generated image is a conceptual blend of the two prompts. From our
findings, \PRO{} usually produces inconsistent results, although the
conditioning embedding given to the pipeline always remains the same, the
balance between visual and semantic blending changes drastically from one seed
to another. An instance of this behavior can be observed in its \textit{Kung
fu-Panda} sample at Figure \ref{fig:survey-sample-images}. In this case, the
model generated possibly the best visual blend out of the four methods, however,
out of all the other seeds tested, no other sample was able to achieve the same
level of blending.

As mentioned in Section~\ref{sec:switch}, results from \SWI{} vary considerably
depending on the timestep at which the prompt is switched, finding the right
timestep is crucial to achieve a good visual blend. This is evident in the case
\textit{Tea-Pot} and \textit{Butter-Fly}, as shown in Figure
\ref{fig:comparison}: the images generated from the prompts \textit{Butter} and
\textit{Fly} are visually distant even though both of them start from the same
initial noise. As a result, when in the middle of the diffusion process the
prompt is switched, the model is unable to shift and correct the existing
distribution towards the one of the new prompt and only the first prompt is
retained in the blend.

For this same reason another undesired behavior of \SWI{} is the
\textit{cartoonification} of the produced blend. The diffusion pipeline, when
unable to shift the pixel distribution towards the new prompt, corrects the
existing noisy image latent by progressively removing the high-frequency
details, resulting in a cartoonish image. This behavior can be clearly observed
in the \textit{Kung fu-Panda} blend produced by \SWI{} in Figure
\ref{fig:comparison}. From our experimental results, this behavior does not
affect the other methods.

The \ALT{} method, which alternates between the two prompts at each timestep,
tends to produce consistent results when the two blended concepts are visually
very different. What is arguably even more interesting is the type of visual
blend that this technique produces when the two concepts are both visually and
semantically very different. This is the case of \textit{Tea-Pot} and
\textit{Butter-Fly}, where the model creates an image that literally and
spatially contains both the first and the second prompt. This is also evident in
the \textit{Bull-Pit} blend in Figure \ref{fig:survey-sample-images}, where the
\ALT{} generates what could be described as \textit{a bull in a pit}. \PRO{}
also seems to produce a similar results but once again, it is too inconsistent
across the seeds space to state it as a general rule.

Compared to the other approaches, the \UNE{} method which encodes in the U-Net
the image latent conditioned with the first prompt and then decodes it with the
second one, produces more subtle blends but generally consistent results. This
might be the reason why this is the blend method that performs worse in the
survey, as the visual blend is not as evident as in the other methods.
Particularly interesting its result on the \textit{Kung fu-Pand} blend, \UNE{}
seems to slightly change the visual representation of the first prompt while
matching the colors of the second one. This subtle blending is also evident in
the \textit{Bull-Pit} blend of Figure \ref{fig:survey-sample-images}, where
surprisingly the pipeline creates an image that somewhat resembles a
\textit{pitbull}.

The results of the survey summarized in Table \ref{tab:results} show that the
most preferred method is \SWI{}, however, this comes with some considerations.
In order to better represent each method, in the survey we have chosen the best
settings for each method, in the case of \SWI{} this translates into using the
optimal timestep in which to switch the prompt for each blend. Finding this
value is a tedious process made by trial and error, with no clear and empirical
way to determine it. Although \UNE{} ranked the lowest in the survey, while
comparing its results with the ones of \SWI{} with a fixed switch timestep in
the middle of the diffusion process (Figure \ref{fig:comparison}), it is evident
that the visual blend produced by these two methods are generally similar if not
better in the case of \UNE.


\section{Conclusions}
\label{sec:conclusion}

Through this paper we tried to answer a novel research question: \emph{is it possible to produce visual concept blends through diffusion models?} We compared different possible solutions to force a diffusion model (more specifically Stable Diffusion \cite{rombach2022high}) to generate contents that represent the blend of two separated concepts.
We collected three different alternatives from exiting publications and from the web. Additionally, we propose a completely new method, which we call \emph{\UNE} that exploits the internal architecture of the adopted diffusion model. We collected the outputs of the different methods on 4 different categories of test; namely, \emph{pairs of animals}, \emph{animal and object}, \emph{compound words}, and \emph{real life scenarios}. For each of these categories we produced various different pairs of concepts, and generated all blends (in total, four blended images for each pair of prompts). 

The quality of a blend, as any creative endeavor, has a subjective 
component on it. Thus, to evaluate which approach is more adept at this
task (in relation to human perception) we devised a user study that was
run by 23 subjects. In it, participants were asked to rank the results
of the blending methods. It is worth noting that two participants did
not rank all methods, but 21 full surveys were submitted. We still used
the partial surveys to compare those pairs where the ranks were available.

From the user study it results that there is no single best blending
method, but the perceived quality varies from pair to pair and, more
importantly, from category to category. And yet, from a positive
perspective, we can answer our research question on the affirmative:
is it possible to produce visual conceptual blends through diffusion
models, and the results are often quite compelling (see the
samples in Figure~\ref{fig:survey-sample-images}. Indeed, the survey
participants expressed surprise with some of them.

An important point to make is that, for this work, we used the
latent space from Stable Diffusion directly; that is, without any
kind of fine-tuning or added training. Thus, our results are less
fragile towards model updates, and do not require significant effort
to implement and execute. This is consistent with our original stated
goal of understanding how to manipulate the latent space as a representation of concepts.
This work only scratches the surface of this topic and we hope that it can inspire new discussion and further analysis. 

For future work, note that our blends are based on very simple 
(mainly one-worded) prompts. This allows us to better understand the
impact of the operations (in contrast to the subtleties of prompt-engineering) but has the disadvantage of working over very general 
concepts, in general, and more in particular is prone to 
ambiguities and misinterpretations. It would thus be interesting to
explore ways to guarantee a more specific identification of concepts
selected for blending.

\begin{acknowledgments}
	This work was partially supported by the MUR for the Department of
	Excellence DISCo at the University of Milano-Bicocca and under the PRIN project
	PINPOINT Prot.\ 2020FNEB27, CUP H45E21000210001; by the FAIR Foundation through
	the project AMAR; and by the NVIDIA Corporation with the RTX A5000 GPUs granted
	through the Academic Hardware Grant Program to the University of Milano-Bicocca
	for the project ``Learned representations for implicit binary operations on
	real-world 2D-3D data.''
\end{acknowledgments}

\bibliography{mainBLEND24}

\end{document}